\documentclass[journal,10pt,twocolumn,twoside]{IEEEtran}

\usepackage[utf8]{inputenc}
\usepackage[T1]{fontenc}
\usepackage{lmodern}
\usepackage[english]{babel}

\usepackage{amsmath,amssymb,amsfonts}
\usepackage{mathrsfs}

\usepackage{graphicx}
\usepackage{epstopdf}
\usepackage{subcaption}
\usepackage[figuresright]{rotating}

\usepackage{booktabs}

\usepackage{algorithm}
\usepackage[noend]{algpseudocode}
\usepackage{orcidlink}
\usepackage{xcolor}
\usepackage{hyperref}
\hypersetup{colorlinks=true,linkcolor=blue,urlcolor=blue,citecolor=blue}
\usepackage{cite}

\DeclareMathOperator*{\argmax}{arg\,max}
\newcommand{\mat}[1]{\mathbf{#1}}

\begin{document}

\title{Rainbow Deep Q-Learning with Kinematics-Aware Design for Cooperative Delta and 3-RRS Parallel Robot Insertion}

\author{%
    Hassen Nigatu\,\orcidlink{0000-0002-1825-0097},
    Shi Gaokun\,\orcidlink{0000-0002-4692-0115},
    Li Jituo\,\orcidlink{0000-0003-1343-5305},
    Wang Jin\,\orcidlink{0000-0003-3106-021X},
    Lu Guodong\,\orcidlink{0009-0004-8833-5071},
    and Howard Li%
    \thanks{Manuscript received Month DD, 2026; revised Month DD, 2026.
        This work was supported in part by the Robotics Research Center
        of Yuyao under Grant KZ22308, in part by the Ningbo Yongjiang
        Talent Program under Grant Z22501, in part by the Zhejiang
        Talents Program, and in part by the National Natural Science
        Foundation of China under Grant 52275276.
        \textit{(Corresponding authors: Li Jituo; Howard Li.)}}%
    \thanks{H. Nigatu, S. Gaokun, J. Li, J. Wang, and G. Lu are with the
        Robotics Institute of Zhejiang University (Yuyao Robotics
        Research Centre), Yuyao Technology Innovation Center, No.~479,
        Yuyao, Ningbo 315400, Zhejiang, China (e-mail:
        hassen@zju.edu.cn; gaokun@zju.edu.cn; jituo\_li@zju.edu.cn;
        wangjin@zju.edu.cn; luguodong@zju.edu.cn).}%
    \thanks{H. Li is with the Department of Electrical and Computer
        Engineering, University of New Brunswick, Fredericton, NB
        E3B~5A3, Canada (e-mail: howard@unb.ca).}%
    \thanks{Digital Object Identifier 10.1109/TRO.2026.XXXXXXX}%
}

\markboth{IEEE Transactions on Robotics,~Vol.~XX, No.~X, Month~2026}%
{Nigatu \MakeLowercase{\textit{et al.}}: Rainbow Deep Q-Learning with Kinematics-Aware Design for Cooperative Delta and 3-RRS Insertion}

\maketitle

\begin{abstract}
This paper presents a kinematics-aware deep reinforcement learning
framework based on Rainbow Deep Q-Networks (DQN) for cooperative
peg-in-hole manipulation by a Delta parallel robot and a 3-RRS
(Revolute--Revolute--Spherical) parallel manipulator. A key contribution
is the integration of a geometric design-optimization stage that
precedes learning: the 3-RRS geometry is tuned to maximize the
singularity-free workspace and improve conditioning, which in turn
enlarges the safe region in which the reinforcement learning policy
can explore. Together the two manipulators expose a 6~degree-of-freedom
(DoF) controllable subspace (three Delta translations, two 3-RRS
rotations, and one 3-RRS vertical translation); the peg-in-hole task is
invariant to rotation about the peg axis, so the task-relevant manifold
is five dimensional. The cooperative insertion problem is cast as a
Markov Decision Process with a 12-dimensional state vector and a
discrete action set containing $6 \times 2 = 12$ incremental commands
(one positive and one negative per controlled DoF). A shaped reward
combines dense proximity guidance, penalties for kinematic and
workspace violations, and sparse bonuses for successful insertions.
The Rainbow DQN---integrating double Q-learning, dueling architecture,
prioritized replay, multi-step returns, noisy linear layers for
exploration, and a distributional value head---is trained with a
two-stage curriculum. The co-designed framework is validated in a
high-fidelity kinematic simulator, where it achieves stable policy
convergence, reliable insertions, and reduced constraint violations
compared against a vanilla DQN agent and a classical sampling-based
planner.
\end{abstract}


\IEEEpeerreviewmaketitle

\section{Introduction}\label{sec:introduction}
Cooperative manipulation by multiple robots is a central research
direction in modern robotics, driven by demand for high-precision
operations in dynamic, unstructured environments. Applications range
from industrial assembly and surgical automation to search-and-rescue
and collaborative handling, all of which rely on tight spatial and
temporal coordination among agents~\cite{SuttonBarto2018,
Chen2025MRSsoft}. Unlike single-robot systems, cooperative teams must
continuously resolve inter-agent dependencies, negotiate shared
workspaces, and adapt to evolving task requirements under timing
constraints~\cite{Sun2023PegSurvey}.

\begin{figure}[t]
    \centering
    \includegraphics[width=\columnwidth]{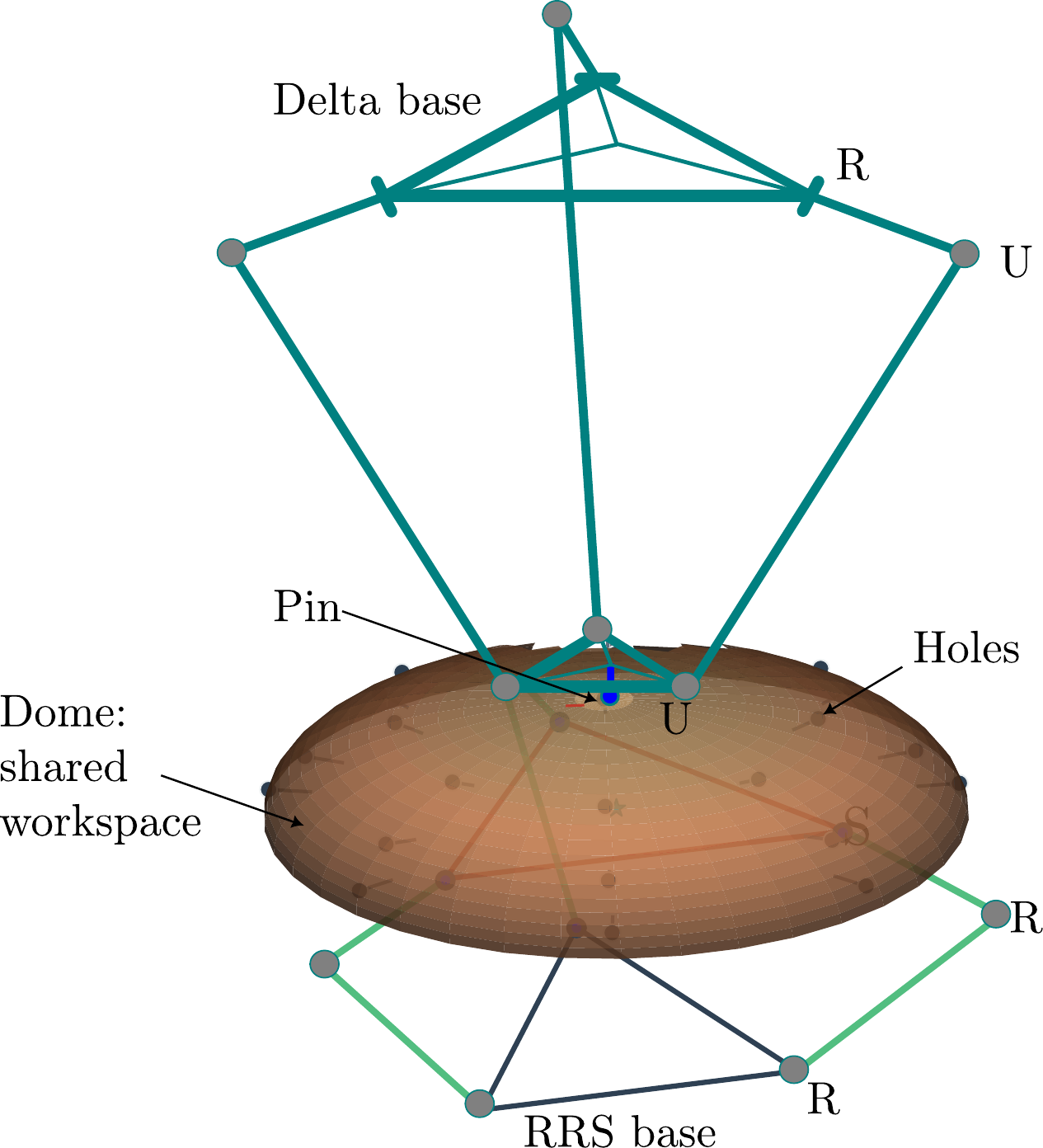}
    \caption{Schematic of the Delta and 3-RRS parallel manipulators
    in a cooperative configuration, with a dome-shaped shared
    workspace for synchronized peg-in-hole task execution.}
    \label{fig:arch}
\end{figure}

Classical motion planning and control are effective in isolated,
structured settings but face scalability and robustness challenges in
cooperative scenarios. Deep reinforcement learning (DRL) has emerged
as a complementary paradigm that enables adaptive, data-driven policy
synthesis~\cite{Mnih2015, Tang2024DRLRealWorld}. Multi-agent DRL
formulations such as centralized training with decentralized
execution~\cite{Lowe2017MADDPG} and decentralized policy
learning~\cite{Keval2023decentralised} have shown promise for dynamic
coordination problems where explicit planning pipelines fall
short~\cite{Mtowe2025low}. In this work we focus on a simpler but
still-effective cooperative design: a \emph{centralized single-agent}
controller that commands both manipulators over a joint action space.
This framing avoids the communication and non-stationarity issues of
fully decentralized multi-agent DRL while still exercising cooperative
behaviour, and is appropriate because the two manipulators share a
common workspace and a single task objective.

We study the cooperative system depicted in
Fig.~\ref{fig:arch}: a Delta parallel manipulator~\cite{Williams2016DeltaKin,
Clavel1990Patent} executes translational positioning of the peg, while
a 3-RRS manipulator~\cite{DiGregorio2004JMD3RRSWrist, Guo2020Robotica3RRS}
orients a dome-shaped platform carrying the target
holes~\cite{Li2001ICRA3RRS, Tetik2017Kinematics3RRS}. The problem
combines three coupled challenges. First, \emph{synchronization}
demands precise temporal alignment of translation and rotation under
dynamic conditions. Although centralized cooperative schemes such as
MADDPG~\cite{Lowe2017MADDPG} have shown coordination capability, they
incur scalability limits with increasing system
complexity~\cite{Ace2025asynchronous}, and communication-based
methods~\cite{Foerster2016DIAL} introduce latency that is at odds with
millimetre-level insertion tolerance~\cite{Chen2024DeltaPeg}. Second,
\emph{adaptation} to changing relative pose between peg tip and target
hole is required without explicit
reprogramming~\cite{Tobin2017DomainRandomization, Bengio2009Curriculum,
Siciliano2024DualArmPeg, Florensa2017RCG}. Third, \emph{safe motion}
in a confined workspace must respect kinematic constraints; parallel
mechanism singularities~\cite{Tetik2018Singularity3RRS} and workspace
boundary effects~\cite{Zhao2022Ankle3RRSExp} complicate matters.
Existing safety frameworks~\cite{Garcia2015SafeRL, Brunke2022SafeLearning,
Ye2025CollisionSurvey} provide foundations but require task-specific
realizations.

Within this context, we first enlarge the safe exploration region by
geometric optimization of the 3-RRS, then train a Rainbow
DQN~\cite{Hessel2018Rainbow} that integrates double
Q-learning~\cite{VanHasselt2016}, dueling
architectures~\cite{Wang2016}, prioritized experience
replay~\cite{Schaul2015,Schaul2016PER}, multi-step returns, noisy
linear layers for exploration~\cite{Fortunato2017}, and a
distributional value head~\cite{Bellemare2017}. The reward function
uses kinematic knowledge of both manipulators---Delta workspace
constraints~\cite{Williams2016DeltaKin} and the optimized 3-RRS
workspace~\cite{Li2001ICRA3RRS}---for dense shaping, combined with
sparse insertion bonuses~\cite{Shanghai2025VIKI} and safety
penalties~\cite{Garg2024SafeMRS, Li2023SafeMARL_AI, Naito2024task}.
Because reward shaping and action masking both rely on analytic
kinematic models, we describe the method as
\emph{kinematics-informed} DRL rather than strictly model-free.

\paragraph*{Contributions}
\begin{enumerate}
    \item A \emph{sequential co-design} workflow in which the 3-RRS
    geometric parameters are optimized first, in order to enlarge the
    singularity-free workspace and improve conditioning, and Rainbow
    DQN is then trained in this enlarged safe region for cooperative
    peg-in-hole insertion with a heterogeneous Delta~+~3-RRS pair.
    \item A task-specific Rainbow DQN formulation with a unified
    12-dimensional state, a well-defined discrete action set of 12
    incremental commands over 6 controlled DoF, and a shaped reward
    that combines dense proximity guidance, workspace-violation
    penalties, and sparse insertion bonuses.
    \item A simulation study with ablations over Rainbow components
    and over the geometric-optimization stage, comparing against a
    vanilla DQN and a classical sampling-based planner baseline.
\end{enumerate}

The remainder of the paper is organized as follows.
Section~\ref{sec:learning_based_control} reviews learning-based control
background. Section~\ref{sec:proposed_method} presents our integrated
approach. Section~\ref{sec:experiment} reports simulation results and
analysis. Section~\ref{sec:discussion} discusses limitations, and
Section~\ref{sec:conclusion} concludes.

\section{Learning-Based Control}\label{sec:learning_based_control}
Reinforcement Learning (RL) is a framework for sequential
decision-making in which an agent interacts with an environment to
learn a policy that maximizes cumulative reward~\cite{SuttonBarto2018}.
At each discrete time step $t$, the agent observes a state
$s_t\in\mathcal{S}$, selects an action $a_t\in\mathcal{A}$ according to
a policy $\pi(a|s)$, and receives a reward $r_t\in\mathbb{R}$. The
environment transitions to $s_{t+1}$ according to the kernel
$P(s_{t+1}\mid s_t,a_t)$. The interaction is modeled as a Markov
Decision Process $(\mathcal{S},\mathcal{A},P,R,\gamma)$ with discount
factor $\gamma\in[0,1]$. The optimal policy maximizes the expected
discounted return
\begin{equation}
    G_t = \mathbb{E}_{\pi}\!\left[\sum_{k=0}^{\infty}\gamma^k r_{t+k}\right].
    \label{eq:discounted_return}
\end{equation}

\subsection{Value-Based, Policy-Based, and Actor--Critic Methods}
Value-based algorithms estimate $V(s)$ or $Q(s,a)$ and act greedily
with respect to the estimate. Policy-based methods parameterize
$\pi_\theta(a|s)$ and optimize the expected return by gradient
ascent, which suits continuous and stochastic policies. Actor--critic
methods combine both. In our setting---discrete incremental actions
and a shared workspace with hard kinematic constraints---value-based
methods are convenient because the finite action set enables
straightforward action masking via inverse kinematics, a property we
rely on throughout Section~\ref{sec:proposed_method}.

\subsection{Q-Learning and the Bellman Optimality Equation}
Q-learning~\cite{Watkins1989,Watkins1992} estimates $Q^*(s,a)$ via the
Bellman optimality equation
\begin{equation}
    Q^*(s,a)=\mathbb{E}\!\left[r_t+\gamma\max_{a'}Q^*(s_{t+1},a')
    \mid s_t=s, a_t=a\right].
    \label{eq:bellman_optimality}
\end{equation}
The tabular update is
\begin{equation}
    Q(s_t,a_t)\!\leftarrow\!Q(s_t,a_t)+\alpha\!\left[r_t+\gamma\max_{a'}
    Q(s_{t+1},a')-Q(s_t,a_t)\right].
    \label{eq:qlearning_update}
\end{equation}
Tabular Q-learning does not scale to the high-dimensional joint
workspace of coordinated Delta~+~3-RRS control, motivating function
approximation.

\subsection{Deep Q-Network (DQN)}
DQN~\cite{Mnih2015} approximates $Q_\theta(s,a)$ with a neural network
trained to minimize the temporal-difference loss
\begin{equation}
    L_t(\theta)=\mathbb{E}_{(s,a,r,s')\sim\mathcal{D}}\!\left[
    \left(y_t-Q(s,a;\theta)\right)^2\right],
    \label{eq:dqn_loss}
\end{equation}
with target
\begin{equation}
    y_t=r+\gamma\max_{a'}Q(s',a';\theta^{-}),
    \label{eq:dqn_target}
\end{equation}
replay buffer $\mathcal{D}$, and periodically updated target
parameters $\theta^-$. Vanilla DQN suffers from overestimation bias,
slow reward propagation, and poor exploration under sparse rewards.

\subsection{Rainbow DQN}\label{subsec:rainbow_dqn}
Rainbow DQN~\cite{Hessel2018Rainbow} integrates six extensions of the
original DQN: double Q-learning~\cite{VanHasselt2016} to reduce
overestimation bias; the dueling architecture~\cite{Wang2016} for
separate state-value and advantage estimation; prioritized experience
replay~\cite{Schaul2015,Schaul2016PER} for sample efficiency; multi-step
returns for faster reward propagation; NoisyNets~\cite{Fortunato2017}
for state-dependent exploration; and distributional RL~\cite{Bellemare2017}
for richer value representation. These extensions directly address the
limitations that are most problematic in long-horizon, sparse-reward
coordinated manipulation: sample inefficiency, overestimation bias,
and exploration. This motivates our choice of Rainbow DQN for the
discrete-action formulation in Section~\ref{sec:proposed_method}.=
\section{Proposed Method}\label{sec:proposed_method}
We combine geometric optimization of the 3-RRS with Rainbow DQN-based
control. Section~\ref{subsec:problem_definition} formulates the
cooperative insertion task as an MDP. Section~\ref{subsec:optimization}
presents the geometric optimization stage.
Section~\ref{subsec:DQL_approach} details the Rainbow DQN
implementation.

\subsection{Problem Formulation}\label{subsec:problem_definition}
We formalize the cooperative peg-in-hole task as a finite-horizon
Markov Decision Process
\begin{equation}
    \mathcal{M}=\bigl(\mathcal{S},\mathcal{A},\mathcal{T},
    \mathcal{R},\gamma,T_{\max}\bigr),
\end{equation}
with $\gamma\in(0,1)$ and episode horizon $T_{\max}$. The observations
are treated as full state (perfect kinematic sensing in simulation).
A single centralized policy controls both manipulators via a discrete
action set (Algorithm~\ref{alg:Rainbow_DQN}).

\subsubsection{Continuous Configuration Spaces}

\paragraph{Delta subsystem}
Let $\mathbf{p}_{\Delta}(t)=[x_{\Delta}(t),y_{\Delta}(t),z_{\Delta}(t)]^{\top}
\in\mathbb{R}^{3}$ denote the Cartesian pose of the moving platform
centre in the global frame. With active-rod length $l_{a}$ and
passive-rod length $l_{p}$, the admissible workspace is
\begin{equation}
    \begin{aligned}
    \mathcal{W}_{\Delta}=\bigl\{(x,y,z)\in\mathbb{R}^{3}\mid\,
    &z\in[-(l_{a}+l_{p}),-(l_{a}-l_{p})],\\
    &\sqrt{x^{2}+y^{2}}\le r_{\max}\bigr\},\label{eq:delta-ws}
    \end{aligned}
\end{equation}
with a conservative planar bound $r_{\max}=0.8\min\{l_{a},l_{p}\}$.
The inverse kinematic mapping
$\mathcal{I}_{\Delta}:\mathcal{W}_{\Delta}\to\mathbb{R}^{3}$ returns
active-joint angles
$\boldsymbol{\phi}=[\phi_{1},\phi_{2},\phi_{3}]^{\top}$ satisfying
\begin{equation}
    \bigl\|\mathbf{A}_{i}(\boldsymbol{\phi})-\mathbf{P}_{i}(\mathbf{p}_{\Delta})
    \bigr\|_{2}=l_{p},\quad i=1,2,3.
    \label{eq:delta-ik}
\end{equation}
A configuration is \emph{valid} iff~\eqref{eq:delta-ik} admits a real
solution.

\paragraph{3-RRS subsystem}
Let the 3-RRS platform orientation be parameterized by roll--pitch
angles
\(
\boldsymbol{\eta}(t)=[\theta_{\mathrm{roll}}(t),\theta_{\mathrm{pitch}}(t)]^{\top}
\)
and vertical displacement $z_R(t)$. The reachable set is
\begin{equation}
    \mathcal{W}_{\mathrm{RRS}}=\bigl\{(\boldsymbol{\eta},z_R)\mid
    z_R\in[h_{\min},h_{\max}],\;\|\boldsymbol{\eta}\|_{\infty}\le\pi/4\bigr\},
\end{equation}
with $h_{\min},h_{\max}$ set at initialization. Platform attachment
points follow from
$\mathbf{R}(\boldsymbol{\eta})=R_{x}(\theta_{\mathrm{roll}})R_{y}(\theta_{\mathrm{pitch}})$
and $\mathbf{d}=[0,0,z_R]^{\top}$. A 3-RRS configuration is valid if
each limb length satisfies
$|\,\|\mathbf{J}_{i}\mathbf{B}_{i}\|_{2}-L_{1}|\le L_{2}$.

\subsubsection{State Space}
At each time step $t$ the agent receives
\begin{equation}
    s_t=\bigl[\underbrace{\mathbf{p}_{\Delta}(t)}_{3},\;
    \underbrace{\boldsymbol{\eta}(t),z_R(t)}_{3},\;
    \underbrace{\mathbf{e}_{\mathrm{rel}}(t)}_{3},\;
    \underbrace{\mathbf{n}_{\mathrm{target}}(t)}_{3}\bigr]\in\mathbb{R}^{12},
    \label{eq:state}
\end{equation}
where $\mathbf{e}_{\mathrm{rel}}(t)=\mathbf{h}_{\mathrm{target}}-\mathbf{p}_{\mathrm{pin}}$
is the peg-tip to target-hole displacement and
$\mathbf{n}_{\mathrm{target}}$ is the outward hole normal. The state
thus encodes the five task-relevant DoF directly: three Delta
translations, two rotations of the 3-RRS platform, and the vertical
height of the 3-RRS. The peg-in-hole task is invariant to rotation
about the peg axis, so we deliberately do not parameterize that
additional DoF.

\subsubsection{Action Space}
Six DoF are controlled: three Delta translations
$(x_{\Delta},y_{\Delta},z_{\Delta})$, two 3-RRS rotations
$(\theta_{\mathrm{roll}},\theta_{\mathrm{pitch}})$, and the 3-RRS
vertical displacement $z_R$. The discrete action set contains one
positive and one negative increment per DoF, giving
$|\mathcal{A}|=6\times 2=12$. Translational increments use
$\delta_{\mathrm{pos}}=\pm0.02\,\mathrm{m}$; rotational increments use
$\delta_{\mathrm{rot}}=\pm0.03\,\mathrm{rad}$. The two magnitudes
arise from the unit difference between translations and rotations and
were chosen empirically: larger values overshoot during fine
approach, while smaller values lengthen episodes beyond practical
training budgets. After each action, validity is checked against
$\mathcal{W}_{\Delta}\times\mathcal{W}_{\mathrm{RRS}}$ via the analytic
inverse kinematics; invalid actions are masked during action selection.

\subsubsection{Reward Design}\label{subsubsec:reward}
Let
\begin{align}
    d(t)&=\|\mathbf{e}_{\mathrm{rel}}(t)\|_{2}, \\
    \Delta d(t)&=d(t-1)-d(t), \\
    [x]_{+}&=\max(x,0).
\end{align}
Let $v_t\!\in\!\{0,1\}$ indicate a workspace or IK violation induced by
the chosen action; $z_t\!\in\!\{0,1\}$ indicate a \emph{new} successful
insertion at time $t$ (a hole not previously filled); $u_t\!\in\!\{0,1\}$
indicate an attempted duplicate insertion into a hole already filled.
Let $N_t$ denote the total number of holes filled at time $t$ (after
updating), and let $T_{\mathrm{task}}\!=\!60\,\mathrm{s}$ be the
reference task duration used to reward fast completion. The shaped
reward is
\begin{equation}
\begin{aligned}
r_t=\;&-3\,v_t\\
&+\bigl[150+25\,N_t+80\,(1-t/T_{\mathrm{task}})\bigr]\,z_t\\
&-1\cdot u_t\\
&+\bigl(1-v_t-z_t-u_t\bigr)\bigl(
-0.01
-d(t)\\
&\quad+50\,[\Delta d(t)]_{+}
+200\,[0.03-d(t)]_{+}\bigr).
\end{aligned}
\label{eq:reward}
\end{equation}
The first row penalizes constraint violations; the second rewards a new
insertion with a base bonus plus a term proportional to the number of
completed insertions and a time bonus; the third discourages repeated
attempts at already-filled holes; the last row, active only in nominal
transitions, provides dense guidance through a small step cost, a
distance shaping term, an incremental progress bonus
$50\,[\Delta d]_{+}$, and a near-target bonus $200\,[0.03-d]_{+}$ that
dominates once the peg is within $3\,\mathrm{cm}$ of the hole. Sparse
insertion bonuses dominate to encourage task completion, while the
dense terms provide local gradient for exploration.

\subsubsection{Curriculum Learning}
Training uses a two-stage curriculum: stage $\mathcal{C}_0$ includes the
four peripheral holes, and stage $\mathcal{C}_1$ includes all six
holes. We switch from $\mathcal{C}_0$ to $\mathcal{C}_1$ once the
moving-average success rate over the last 20 episodes exceeds $75\%$.
This threshold was chosen empirically to ensure competence on the
simpler task before progression. Preliminary experiments with a
three-stage curriculum yielded only marginal further gains.

\subsubsection{Learning Algorithm Overview}
The policy $\pi_\theta:\mathcal{S}\to\mathcal{A}$ is learned with
Rainbow DQN (Algorithm~\ref{alg:Rainbow_DQN}), combining double
Q-learning, a dueling head, prioritized replay, 3-step returns, noisy
linear layers, and a distributional value output. Training uses a
replay buffer of capacity $10^{6}$ and soft target updates with
$\tau=10^{-3}$ (i.e.\ Polyak averaging); no periodic hard copy is
used. Exploration is driven exclusively by the NoisyNet layers, which
remove the need for a separate $\varepsilon$-greedy schedule.

\begin{algorithm}[t]
    \caption{Rainbow DQN for Cooperative Insertion}
    \label{alg:Rainbow_DQN}
    \begin{algorithmic}[1]
        \State \textbf{Init:} Dueling noisy $Q_\theta$; target $Q_{\theta^-}\!\gets Q_\theta$;
            PER buffer $\mathcal{D}$ (capacity $10^6$); LR scheduler; curriculum stage $\mathcal{C}_0$
        \For{episode $=1,2,\dots$}
            \State Reset env; observe $s_0$; resample NoisyNet noise; $\mathcal{Q}_n\gets\varnothing$
            \While{not terminal}
                \State Compute masked $Q_\theta(s,\cdot)$ (invalid actions $\to-\infty$);
                       $a\gets\argmax_a Q_\theta(s,a)$
                \State Step env $\to(r,s',d)$; push $(s,a,r,s',d)$ into $\mathcal{Q}_n$
                \If{$|\mathcal{Q}_n|\ge n$ \textbf{or} $d$}
                    \State Form $n$-step transition $(\tilde s,\tilde a,\tilde r,\tilde s',\tilde d)$
                           with $\tilde r=\sum_{i=0}^{n-1}\gamma^i r_{t-i}$; insert into $\mathcal{D}$
                           with priority $p\!=\!p_{\max}$
                \EndIf
                \If{ready to train}
                    \State Sample batch by $P(i)\propto p_i^\alpha$;
                           $w_i\gets (1/(NP(i)))^\beta/\max_j w_j$
                    \State $a'\gets\argmax_a Q_\theta(\tilde s',a)$ (masked) \Comment{Double DQN}
                    \State $y\gets\tilde r+(1-\tilde d)\gamma^{n}Q_{\theta^-}(\tilde s',a')$
                    \State $L\gets\tfrac{1}{B}\sum_i w_i\,\mathrm{Huber}(Q_\theta(\tilde s,\tilde a)-y)$
                    \State Adam step on $\theta$ (lr $10^{-4}$, wd $10^{-5}$, grad clip $5$)
                    \State $p_i\gets|Q_\theta(\tilde s,\tilde a)-y|+\epsilon_{\mathrm{per}}$
                    \State $\theta^-\gets(1-\tau)\theta^-+\tau\theta$, $\tau=10^{-3}$
                \EndIf
                \State Resample NoisyNet noise; $s\gets s'$
                \If{$d$} \State \textbf{break} \EndIf
            \EndWhile
            \State Update success metrics; advance curriculum if threshold met;
                   step LR scheduler
        \EndFor
    \end{algorithmic}
\end{algorithm}

\subsection{Geometric Optimization for Learning-Compatible 3-RRS Design}
\label{subsec:optimization}

To provide a well-conditioned environment for the DRL policy, we first
optimize the 3-RRS geometric parameters. A larger singularity-free
workspace and more uniform manipulability enlarge the safe region in
which the policy can explore, which in preliminary tests produced
visibly smoother training curves; quantitative evidence is reported in
Section~\ref{sec:experiment}.

\subsubsection{Motivation}
A well-conditioned mechanism matters for DRL for three reasons: (i)
it enlarges the valid region of the action masking used in
Algorithm~\ref{alg:Rainbow_DQN}, so fewer candidate actions are
rejected; (ii) more uniform manipulability reduces abrupt changes in
transmission that complicate value learning; (iii) a larger
singularity-free workspace reduces the fraction of episodes that
terminate on violations.

\subsubsection{Computational Framework}\label{sec:computational_framework}

\subsubsection{Algorithmic Pipeline Overview}
The pipeline comprises four components:
\begin{enumerate}
    \item \textbf{Kinematic Modeling.} Analytic forward and inverse
    kinematics for the 3-RRS architecture are used for feasibility
    checks during optimization.
    \item \textbf{Workspace Delineation.} A grid-based sampling of
    $\mathcal{W}_{\mathrm{RRS}}$ quantifies reachable volumes and
    singularity loci. Orientations are sampled on a $1^\circ$ grid over
    $\theta_x,\theta_y\in[-\pi/3,\pi/3]$, chosen as a balance between
    resolution and compute time. At each grid point the Jacobian
    $\mat{J}$ is evaluated; singularities are detected by the minimum
    singular value $\sigma_{\min}(\mat{J})<10^{-6}$, and manipulability
    is assessed via the condition number
    $\kappa=\sigma_{\max}/\sigma_{\min}$.
    \item \textbf{Geometric Optimization.} Dimensionless parameters are
    optimized with a gradient-free simplex (Nelder--Mead) method
    subject to the physical constraints below.
    \item \textbf{Performance Validation.} Test trajectories are
    generated on the optimized design to verify joint excursions and
    smoothness.
\end{enumerate}
All analysis uses dimensionless parameters for scale invariance.

\subsubsection{Optimization Formulation}
With base radius $R_b$, platform radius $R_p$, and proximal-link length
$L_1$, define the scale and dimensionless parameters
\begin{equation}
    \eta=\frac{R_b+2L_1+R_p}{4},\;
    \lambda_1=\frac{R_b}{\eta},\;
    \lambda_2=\frac{2L_1}{\eta},\;
    \lambda_3=\frac{R_p}{\eta},
\end{equation}
subject to
\begin{equation}
    \begin{cases}
    \lambda_1+\lambda_2+\lambda_3=4,\\
    \lambda_3>0,\quad\lambda_3<\lambda_1,\quad\lambda_1<2,\quad\lambda_2>0.
    \end{cases}
    \label{eq:constraints}
\end{equation}
The objective is to maximize the area of the singularity-free
orientation workspace,
\begin{equation}
    \begin{aligned}
        \max_{\lambda_1,\lambda_2}\,A_w&=\iint_{\Omega}d\theta_x\,d\theta_y,\\
        \Omega&=\bigl\{(\theta_x,\theta_y)\mid
        \sigma_{\min}(\mat{J})\ge0.15\bigr\}.
    \end{aligned}
\end{equation}
The threshold $\sigma_{\min}\ge0.15$ was chosen by sensitivity
analysis: lower values admit near-singular configurations during
exploration, while higher values unnecessarily shrink the workspace.
Because $A_w$ is approximated by a grid-based Monte Carlo
integration with jittered sample positions, small run-to-run variation
arises; reported statistics in Table~\ref{tab:results} are averaged
over 10 independent runs.

\subsubsection{Computational Outcomes}
The optimization pipeline yields a 42\% expansion in singularity-free
workspace, from $0.82\,\mathrm{rad}^2$ to $1.16\,\mathrm{rad}^2$, with
$\sigma_{\min}$ improving from $0.08$ to $0.22$
(Table~\ref{tab:results}). Maximum roll/pitch excursions improve to
$\pm 40^{\circ}$, condition-number variation across the workspace is
reduced by $68\%$, and the admissible joint angle range narrows to
$[18.7^\circ,68.3^\circ]$, away from singularity boundaries.

\begin{table*}[t]
    \caption{Geometric Optimization Performance Metrics (mean $\pm$ std over 10 runs).}
    \label{tab:results}
    \centering
    \begin{tabular}{lcc}
        \toprule
        \textbf{Metric} & \textbf{Initial} & \textbf{Optimized} \\
        \midrule
        Orientation workspace (rad$^2$) & $0.82\pm0.03$ & $1.16\pm0.05$ ($+42\%$) \\
        Min $\sigma_{\min}$             & $0.08\pm0.01$ & $0.22\pm0.02$ \\
        Max roll/pitch ($^\circ$)       & $\pm32\pm1$   & $\pm40\pm1$ \\
        Condition number variation (\%) & $38\pm2$      & $12\pm1$ \\
        Joint angle range ($^\circ$)    & $[12,78]$     & $[18.7,68.3]$ \\
        \bottomrule
    \end{tabular}
\end{table*}

\begin{figure}[t]
    \centering
    \begin{subfigure}{0.48\columnwidth}
        \includegraphics[width=\linewidth]{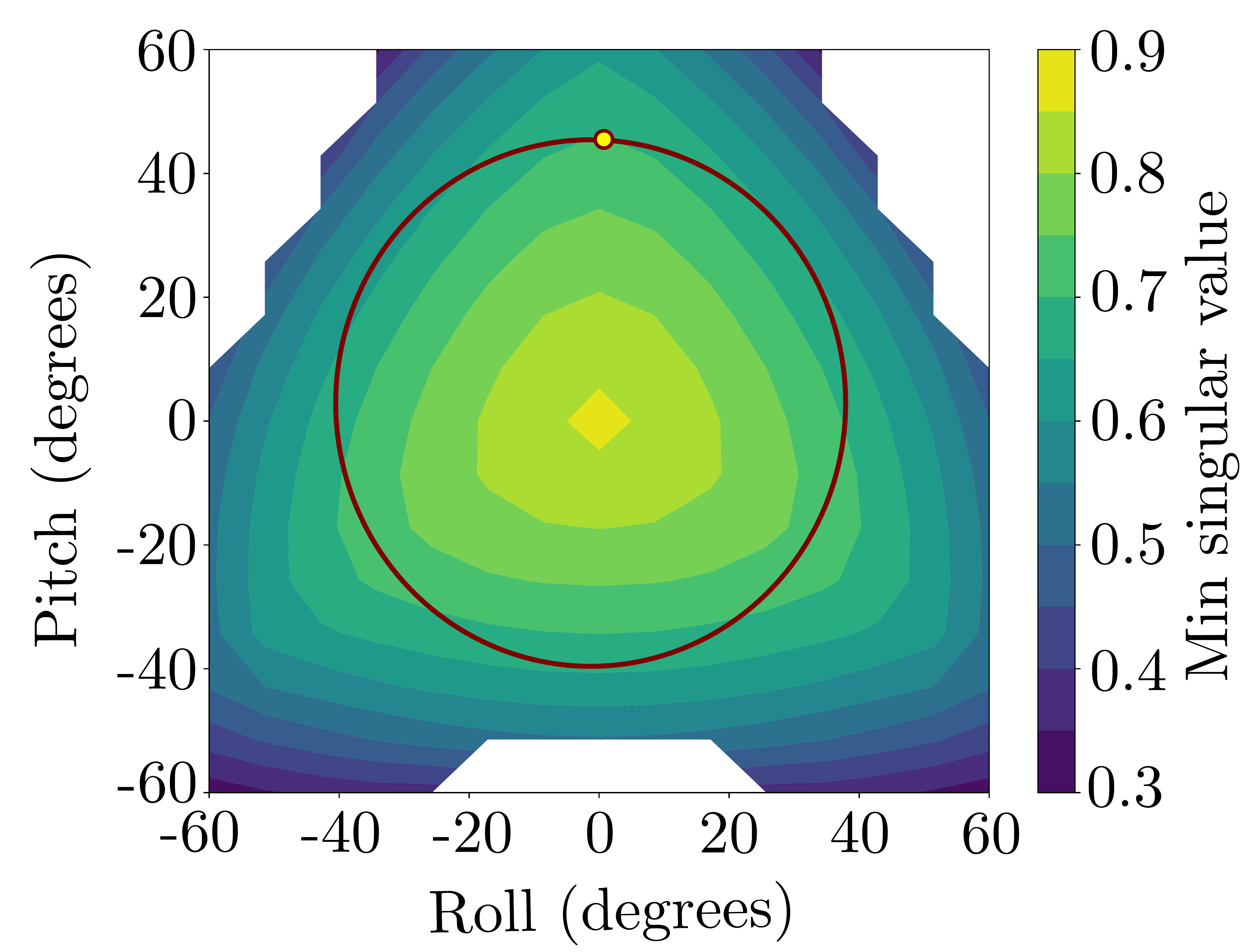}
        \caption{Singularity loci at $z=0.21$\,m}
        \label{fig:singularity_loci}
    \end{subfigure}\hfill
    \begin{subfigure}{0.48\columnwidth}
        \includegraphics[width=\linewidth]{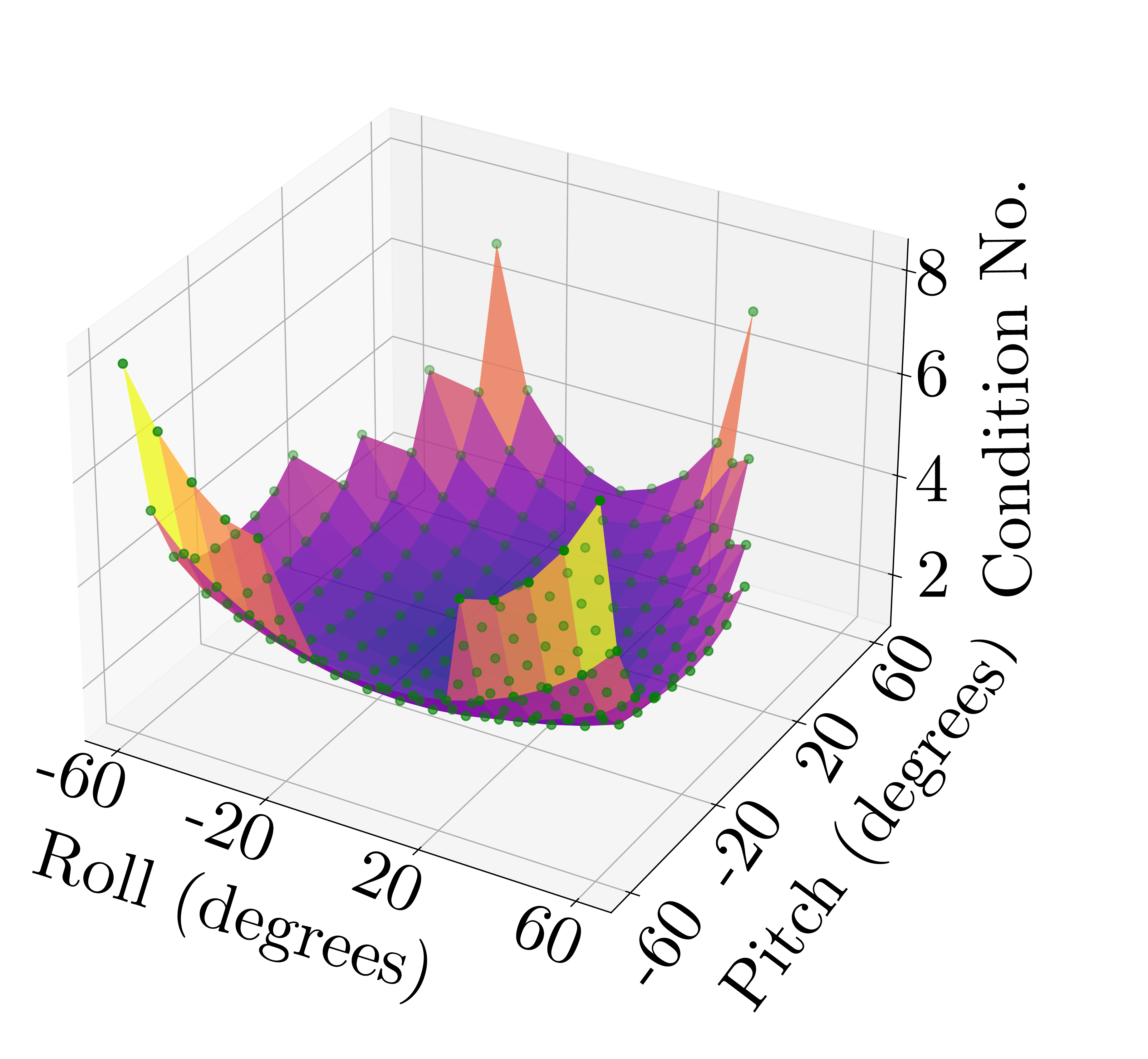}
        \caption{3D manipulability map}
        \label{fig:workspace_3d}
    \end{subfigure}
    \vskip\baselineskip
    \begin{subfigure}{0.48\columnwidth}
        \includegraphics[width=\linewidth]{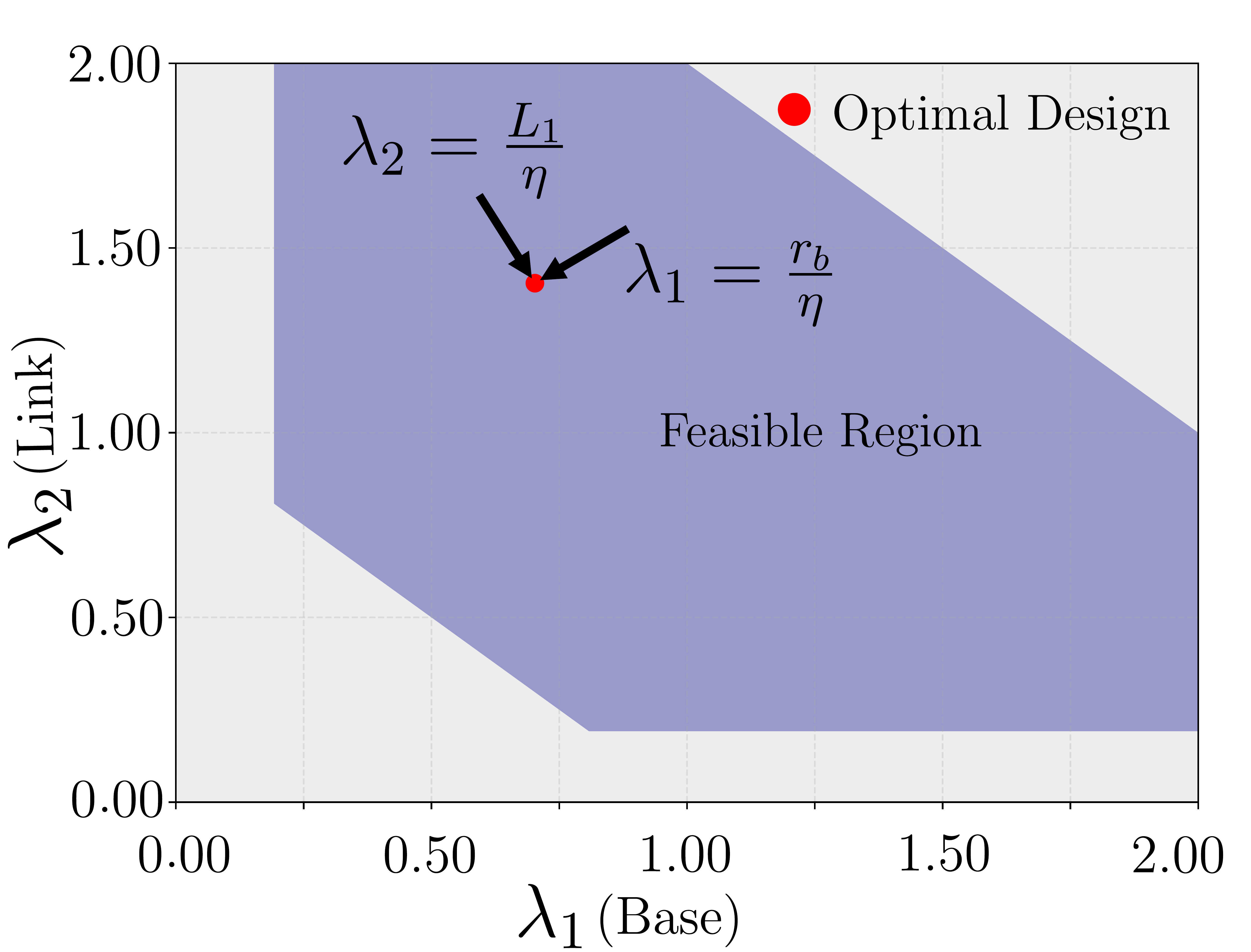}
        \caption{Feasible parameter atlas}
        \label{fig:atlas}
    \end{subfigure}\hfill
    \begin{subfigure}{0.48\columnwidth}
        \includegraphics[width=\linewidth]{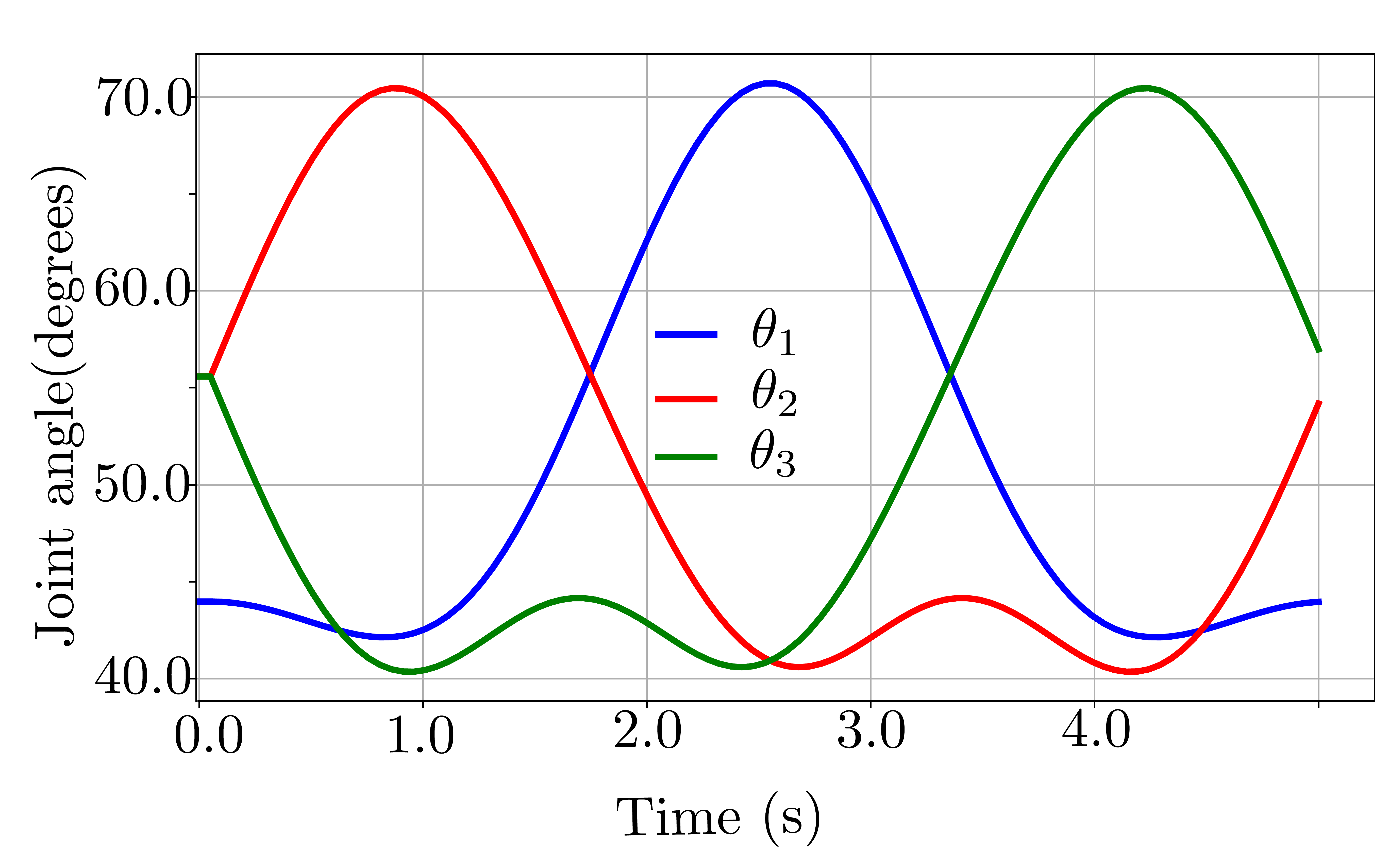}
        \caption{Joint angle trajectories}
        \label{fig:joint_angles}
    \end{subfigure}
    \caption{Analysis outcomes: (a) singularity loci, (b) manipulability
    map, (c) parameter atlas, (d) joint-angle trajectories along a
    representative path.}
    \label{fig:analysis_results}
\end{figure}

Figure~\ref{fig:analysis_results} visualizes the outcomes: the
singularity loci at fixed height, a three-dimensional manipulability
map, the feasible atlas in $(\lambda_1,\lambda_2)$ space, and joint
angle trajectories along a representative path.

\subsubsection{Synergy with the Learning Framework}
To quantify the benefit of the optimization stage on DRL training,
Section~\ref{sec:experiment} reports an explicit ablation that
re-trains the full Rainbow DQN on the non-optimized (initial) 3-RRS
geometry. The improvements reported there are directly attributable
to the enlarged singularity-free workspace, rather than asserted.

\subsection{Rainbow DQN Implementation}\label{subsec:DQL_approach}

\begin{figure*}[t]
    \centering
    \includegraphics[width=0.96\textwidth]{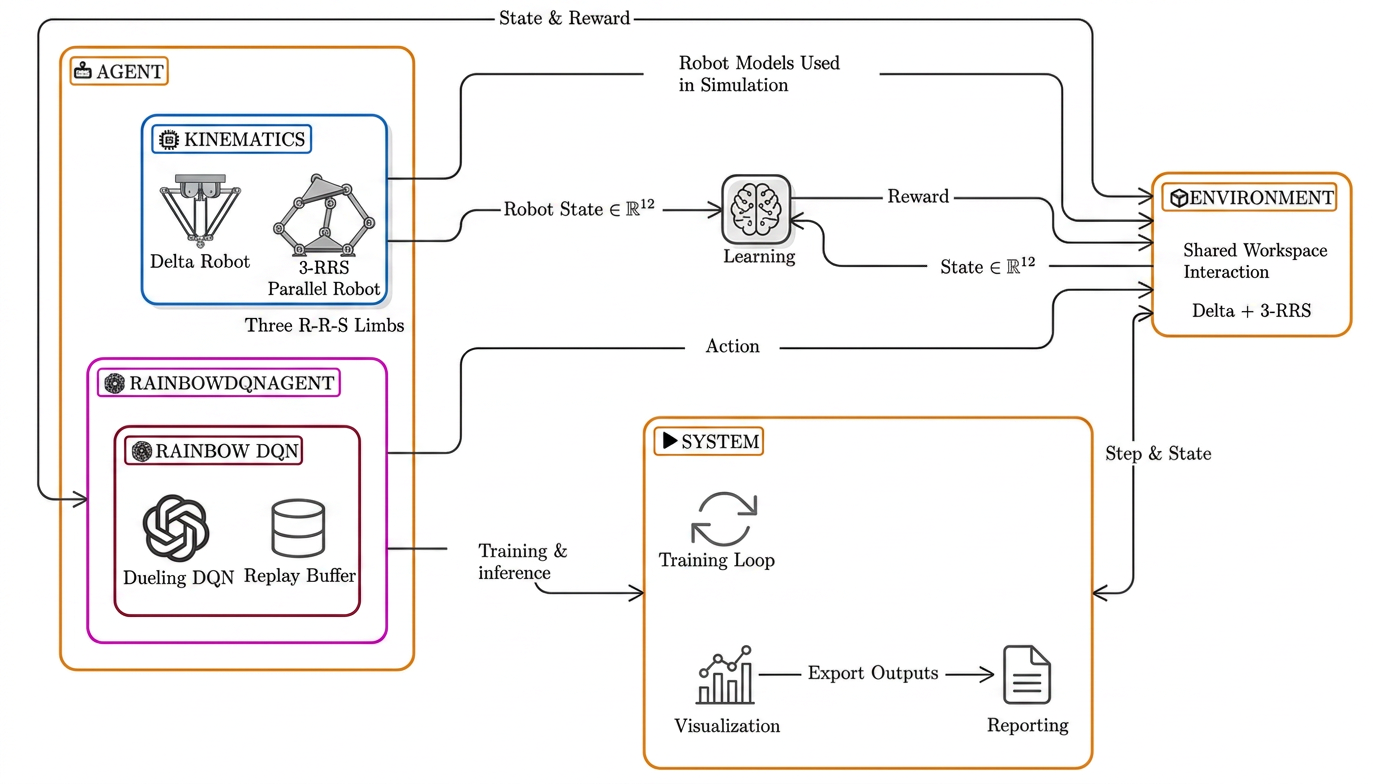}
    \caption{Overall architecture of the Rainbow DQN-based framework.
    The agent integrates kinematic models of the Delta and 3-RRS
    robots with a Rainbow DQN policy. The environment simulates robot
    kinematics and returns states and shaped rewards. Arrows denote
    state, action, and reward signals during end-to-end training for
    cooperative peg-in-hole insertion.}
    \label{fig:dqnarch}
\end{figure*}

Figure~\ref{fig:dqnarch} shows the architecture of our Rainbow
DQN-based framework. The agent block combines analytic kinematic
models of the Delta~\cite{Clavel1990Patent} and
3-RRS~\cite{DiGregorio2004JMD3RRSWrist} mechanisms with a Rainbow DQN
network (dueling head, prioritized replay, noisy linear layers,
$n$-step returns, and a distributional value head). The environment
block simulates robot kinematics and provides state transitions and
shaped rewards; the system block executes the training loop, handles
visualization, and logs metrics.

\subsubsection{State and Action Spaces}
Consistent with Section~\ref{subsec:problem_definition}, the state at
time $t$ is the 12-dimensional vector of~\eqref{eq:state}, combining
$\mathbf{p}_\Delta$, $\boldsymbol{\eta}$, $z_R$, $\mathbf{e}_{\mathrm{rel}}$,
and $\mathbf{n}_{\mathrm{target}}$. The action set contains 12
incremental commands ($\pm0.02\,\mathrm{m}$ for translations,
$\pm0.03\,\mathrm{rad}$ for rotations) over the six controlled DoF.
Discrete actions simplify masking via inverse kinematics; they may
limit trajectory smoothness, which we revisit in
Section~\ref{sec:discussion}.

\subsubsection{Network and Loss}
The action-value function is a distributional, dueling Q-network
$Q(s,a;\theta)$~\cite{Wang2016,Bellemare2017} with noisy linear
layers in the final two layers. The network is trained to minimize
the Huber loss between the predicted atom distribution and the
distributional target built from multi-step returns and the target
network $Q(\cdot;\theta^{-})$:
\begin{equation}
    \label{eq:rainbow_loss}
    \begin{aligned}
        \mathcal{L}_t(\theta)&=\frac{1}{B}\sum_i w_i
        \,\mathrm{Huber}\!\bigl(\mathrm{Proj}(Q(\tilde s,\tilde a;\theta))-y_i\bigr),\\
        y_i&=\mathrm{DistributionalTarget}(\tilde r,\tilde s',\tilde d;\theta^{-}),\\
        \tilde r&=\sum_{i=0}^{n-1}\gamma^i r_{t-i},\\
        w_i&=\bigl(1/(N\,P(i))\bigr)^{\beta},
    \end{aligned}
\end{equation}
with batch size $B$, importance weights $w_i$ from prioritized
replay, discount $\gamma$, and soft target parameters $\theta^-$
updated by Polyak averaging.

\subsubsection{Exploration Strategy}
We rely exclusively on the NoisyNet layers~\cite{Fortunato2017} for
exploration. State-dependent noise is injected into the final dense
layers of the Q-network; as the policy improves, the network learns to
suppress this noise, producing a smooth transition from exploration to
exploitation without a hand-tuned $\varepsilon$-greedy schedule. We
do \emph{not} additionally use $\varepsilon$-greedy.

\section{Simulation Results and Analysis}\label{sec:experiment}

\subsection{Simulation Environment}
We implemented the approach in a custom Python simulator with
deterministic forward and inverse kinematics for both manipulators,
geometric collision detection, and alignment-based insertion success
criteria (position and orientation thresholds). The environment
exposes a unified interface for state observation, action execution,
reward calculation, and episode termination.

The manipulators used are:
\begin{itemize}
    \item \textbf{Delta.} A 3-DoF translational parallel manipulator
    that manipulates the peg and performs the axial insertion motion.
    \item \textbf{3-RRS.} A 3-DoF parallel manipulator providing two
    rotations (roll, pitch) and one vertical translation, carrying a
    dome-shaped platform with six target holes.
\end{itemize}

\subsection{Implementation and Training Protocol}
The Q-network uses three dense hidden blocks (256, 128, 64 units) with
ReLU activation, followed by a dueling distributional head with 51
atoms. The final two dense layers are noisy linear layers. Inputs are
normalized to $[0,1]$ for training stability. We report a single
authoritative hyperparameter set (Table~\ref{tab:hparams}) used for
Algorithm~\ref{alg:Rainbow_DQN} throughout all experiments.

\begin{table}[t]
    \caption{Training hyperparameters (authoritative).}
    \label{tab:hparams}
    \centering
    \begin{tabular}{ll}
        \toprule
        \textbf{Parameter} & \textbf{Value} \\
        \midrule
        Optimizer                     & Adam \\
        Learning rate                 & $10^{-4}$ \\
        Weight decay                  & $10^{-5}$ \\
        Gradient clipping             & $5.0$ \\
        Discount factor $\gamma$      & $0.99$ \\
        $n$-step return               & $3$ \\
        Replay buffer capacity        & $10^{6}$ \\
        PER $\alpha,\,\beta$          & $0.6,\,0.4\to 1.0$ (annealed) \\
        Batch size                    & $64$ \\
        Target update                 & soft, $\tau=10^{-3}$ \\
        Exploration                   & NoisyNets (no $\varepsilon$-greedy) \\
        Distributional atoms          & $51$ in $[V_{\min},V_{\max}]=[-10,200]$ \\
        Total environment steps       & $\sim 1.5\times10^{6}$ \\
        Curriculum switch threshold   & $75\%$ success over 20 episodes \\
        Hardware                      & NVIDIA RTX~3090, $\sim$4\,h training \\
        \bottomrule
    \end{tabular}
\end{table}

Hyperparameters were selected by coarse grid search over learning rate
$\{5\!\times\!10^{-5},10^{-4},3\!\times\!10^{-4}\}$, $n$-step
$\{1,3,5\}$, and batch size $\{32,64,128\}$, evaluated on a validation
protocol of $5$ independent seeds and $100$ post-training episodes
each. Ablations used the same hyperparameters as the full Rainbow.

\subsection{Evaluation Metrics}
\begin{enumerate}
    \item \textbf{Task completion time}: time to successfully insert
    the peg into a target hole.
    \item \textbf{Alignment accuracy}: angular deviation between peg
    axis and hole normal at insertion.
    \item \textbf{Success rate}: fraction of episodes with at least
    one successful insertion (per-episode success).
    \item \textbf{Collision rate}: average collisions per episode.
    \item \textbf{Energy proxy}: integrated joint torque over the
    episode, estimated from joint-space velocity magnitudes.
    \item \textbf{Trajectory error}: RMS deviation from a geometric
    reference path to the target.
\end{enumerate}

\subsection{Baselines}
We compare against two baselines trained or tuned on the same task.

\paragraph{Classical planner (RRT-Connect)}
The baseline is RRT-Connect in OpenRAVE, given the full geometric
model. A centralized coordinator sequences the two manipulators: the
3-RRS is planned to a nominal alignment pose, then the Delta is
planned to execute the insertion, each planning call capped at
$5\,\mathrm{s}$. This baseline has no closed-loop corrective control,
so failure cases are primarily insertion-precision failures rather
than failures to find a feasible path. We report it as a
representative open-loop planning baseline, not as an upper bound.

\paragraph{Vanilla DQN}
The same network architecture and hyperparameters as our approach, but
with none of the six Rainbow extensions (single Q-learning, plain
feed-forward head, uniform replay, 1-step returns,
$\varepsilon$-greedy exploration linearly decayed from $1.0$ to
$0.05$ over $10^{4}$ steps, and scalar value output).

\subsection{Main Results}
Results in Table~\ref{tab:performance} are averaged over 100 test
episodes with 5 seeds. Our Rainbow DQN approach outperforms both
baselines across all metrics (paired $t$-test, $p<0.01$ vs.\ each
baseline).

\begin{table*}[t]
    \caption{Performance comparison between Rainbow DQN, Vanilla DQN, and the classical planner baseline.}
    \label{tab:performance}
    \centering
    \begin{tabular}{lccc}
        \toprule
        \textbf{Metric} & \textbf{Classical planner} & \textbf{Vanilla DQN} & \textbf{Rainbow DQN (Ours)} \\
        \midrule
        Success rate (\%)                & $55\pm4$  & $68\pm5$ & $\mathbf{95\pm2}$ \\
        Avg.\ completion time (s)        & $12.4\pm1.2$ & $9.8\pm0.9$ & $\mathbf{7.1\pm0.5}$ \\
        Avg.\ alignment error (deg)      & $1.8\pm0.3$  & $1.2\pm0.2$ & $\mathbf{0.6\pm0.1}$ \\
        Avg.\ collisions per episode     & $0.5\pm0.1$  & $0.3\pm0.1$ & $\mathbf{0.1\pm0.05}$ \\
        Avg.\ energy proxy (J)           & $15.4\pm1.1$ & $13.2\pm0.8$ & $\mathbf{12.3\pm0.6}$ \\
        RMS trajectory error (mm)        & $4.2\pm0.4$  & $2.8\pm0.3$  & $\mathbf{1.5\pm0.2}$ \\
        \bottomrule
    \end{tabular}
\end{table*}

\begin{figure}[t]
    \centering
    \begin{subfigure}{0.48\columnwidth}
        \includegraphics[width=\linewidth]{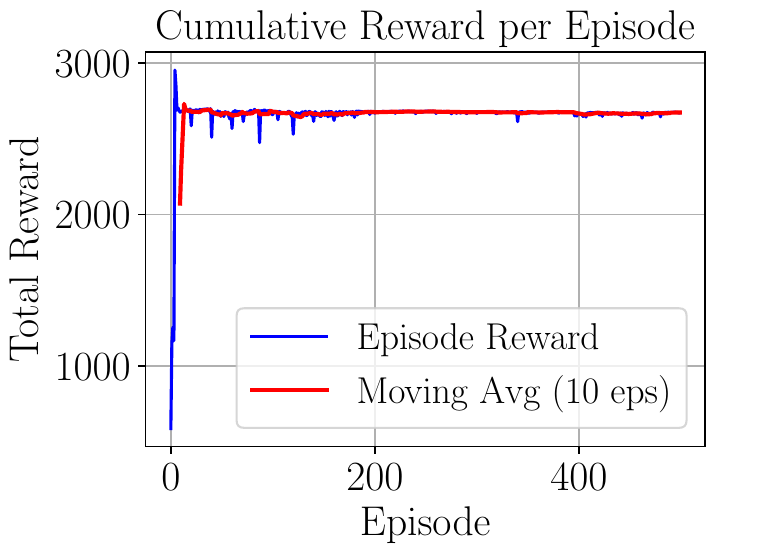}
        \caption{}
        \label{fig:cum_reward}
    \end{subfigure}\hfill
    \begin{subfigure}{0.48\columnwidth}
        \includegraphics[width=\linewidth]{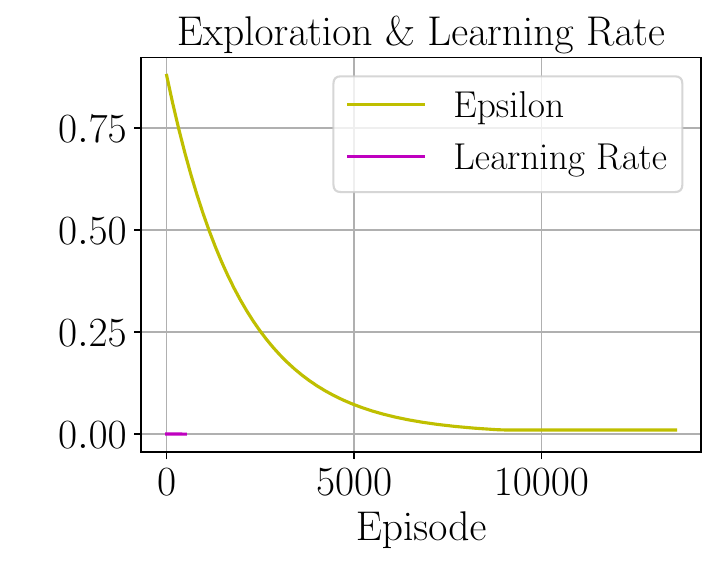}
        \caption{}
        \label{fig:epsilon}
    \end{subfigure}
    \caption{(a) Cumulative reward per episode (raw and 10-episode
    moving average). The step at $\sim$150 episodes marks the
    curriculum switch. (b) Learning-rate annealing and the
    self-regulating NoisyNet noise magnitude over episodes.}
\end{figure}

\begin{figure}[t]
    \centering
    \begin{subfigure}{0.48\columnwidth}
        \includegraphics[width=\linewidth]{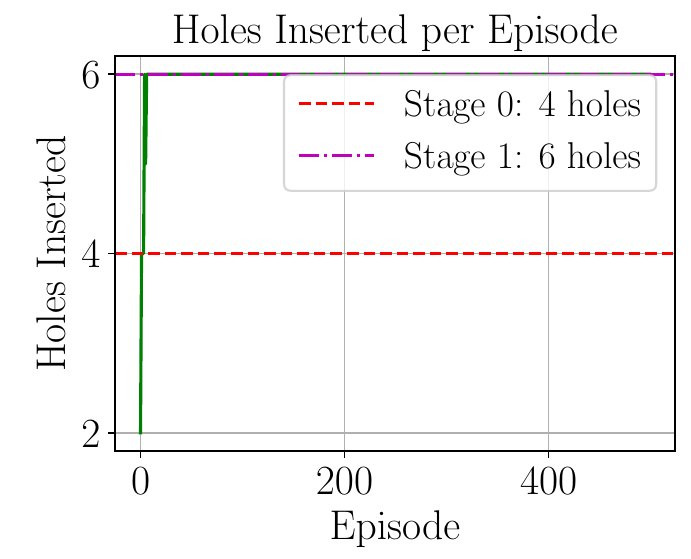}
        \caption{}
        \label{fig:holes}
    \end{subfigure}\hfill
    \begin{subfigure}{0.48\columnwidth}
        \includegraphics[width=\linewidth]{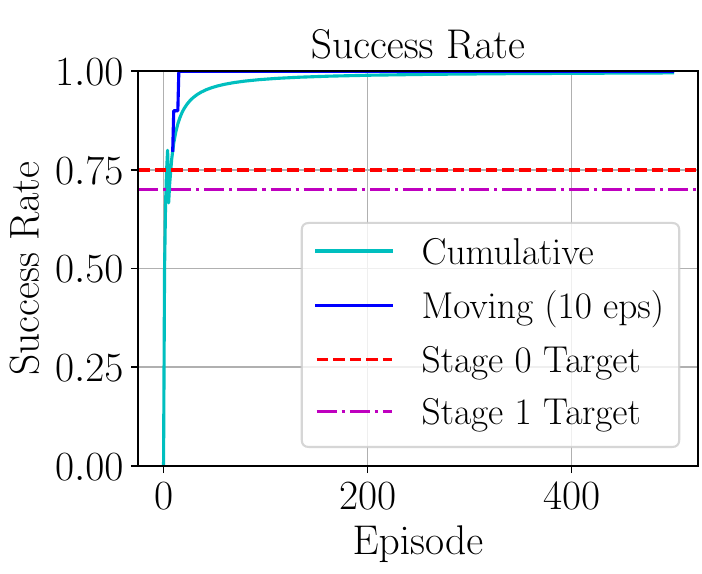}
        \caption{}
        \label{fig:success}
    \end{subfigure}
    \caption{(a) Holes inserted per episode across curriculum stages
    (Stage~0: four holes; Stage~1: six holes). Performance drops
    briefly after the stage switch and recovers. (b) Per-episode
    success-rate trajectory over training.}
\end{figure}

\begin{figure}[t]
    \centering
    \begin{subfigure}{0.48\columnwidth}
        \includegraphics[width=\linewidth]{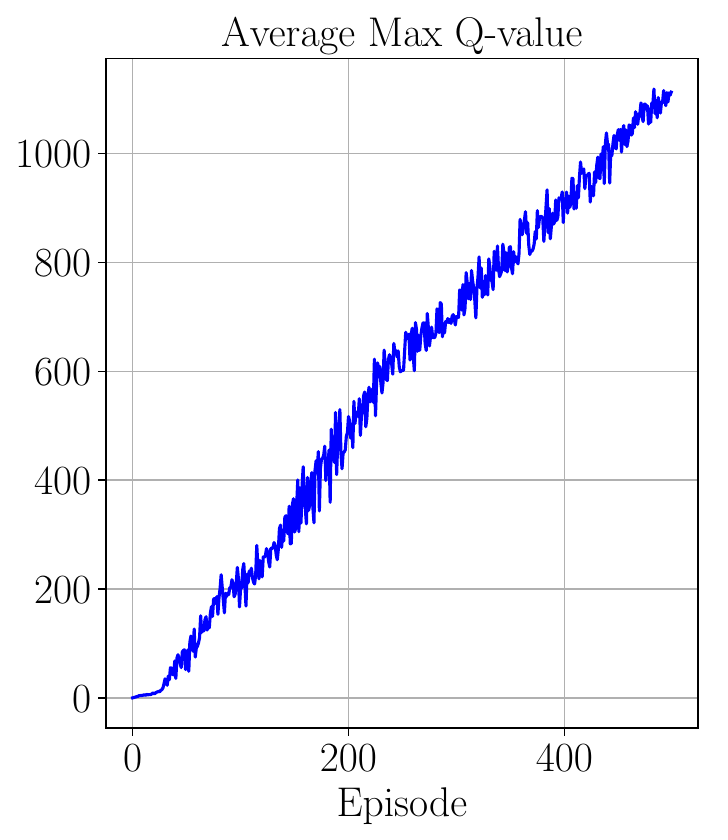}
        \caption{}
        \label{fig:max_q}
    \end{subfigure}\hfill
    \begin{subfigure}{0.48\columnwidth}
        \includegraphics[width=\linewidth]{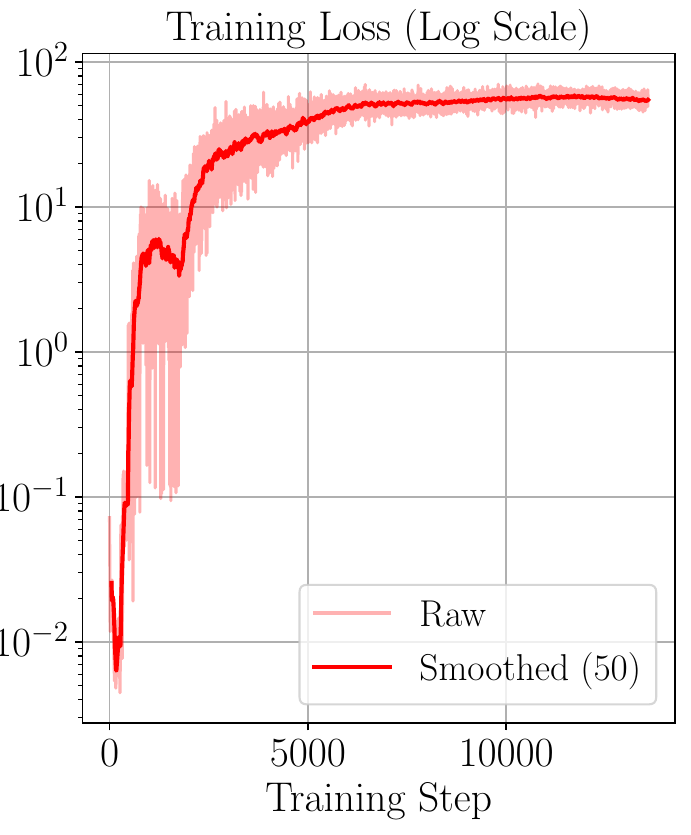}
        \caption{}
        \label{fig:loss}
    \end{subfigure}
    \caption{(a) Average maximum Q-value during training, indicating
    growing agent confidence. (b) Training loss (log scale), showing
    convergence.}
\end{figure}

\begin{figure}[t]
    \centering
    \includegraphics[width=0.9\columnwidth]{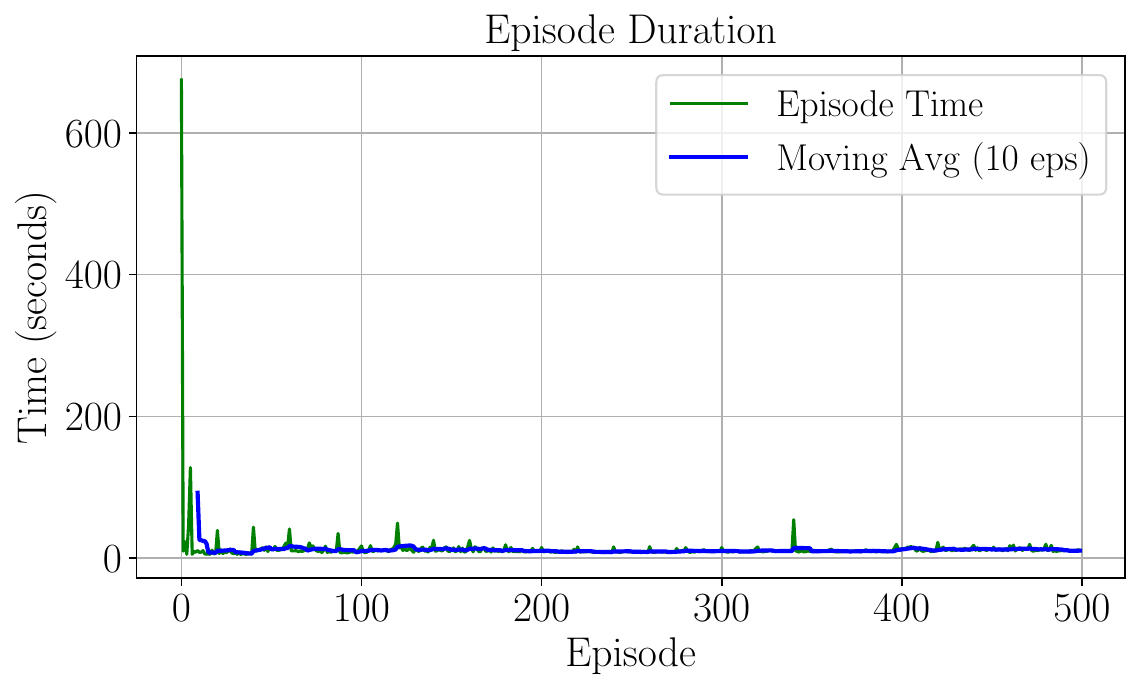}
    \caption{Per-episode duration in seconds with a 10-episode moving
    average. The decreasing trend indicates increasing policy
    efficiency.}
    \label{fig:duration}
\end{figure}

Figure~\ref{fig:cum_reward} shows the learning curves. The two-stage
curriculum produces a brief dip at the stage switch that the policy
quickly recovers from. Figure~\ref{fig:epsilon} reports the
annealing of the learning rate and the NoisyNet noise magnitude, the
latter serving as an effective exploration schedule learned by the
network. Figure~\ref{fig:holes} shows the number of holes inserted
per episode across curriculum stages, and
Figure~\ref{fig:success} the overall success-rate trajectory.
Figure~\ref{fig:max_q} shows the average maximum Q-value, while
Figure~\ref{fig:loss} shows the training loss (log scale).
Figure~\ref{fig:duration} confirms that episode duration decreases
over training as the policy becomes more efficient.

\subsection{Ablation: Rainbow Components}
Table~\ref{tab:ablation} reports success rate when each Rainbow
extension is removed, holding all other hyperparameters fixed and
re-training from scratch. Prioritized replay and the dueling head
contribute most strongly to performance, consistent with
findings in~\cite{Hessel2018Rainbow}.

\begin{table}[t]
    \caption{Ablation: success rate with individual Rainbow components removed.}
    \label{tab:ablation}
    \centering
    \begin{tabular}{lc}
        \toprule
        \textbf{Configuration} & \textbf{Success rate (\%)} \\
        \midrule
        Full Rainbow (ours)        & $\mathbf{95\pm2}$ \\
        $-$ Prioritized replay     & $82\pm3$ \\
        $-$ Dueling head           & $85\pm3$ \\
        $-$ Multi-step returns     & $88\pm2$ \\
        $-$ NoisyNets              & $90\pm2$ \\
        $-$ Distributional head    & $87\pm3$ \\
        $-$ Double Q-learning      & $89\pm2$ \\
        \bottomrule
    \end{tabular}
\end{table}

\subsection{Ablation: Geometric Optimization}
To substantiate the benefit of the geometric optimization stage, we
retrain the full Rainbow DQN on the \emph{non-optimized} initial
3-RRS geometry and compare to the optimized geometry. All other
hyperparameters and training budgets are identical
(Table~\ref{tab:geo_ablation}). The optimized geometry trains faster
(fewer environment steps to $75\%$ success), reaches a higher final
success rate, and encounters far fewer singularity-related
terminations.

\begin{table}[t]
    \caption{Effect of the geometric-optimization stage on Rainbow DQN training (5 seeds, mean $\pm$ std).}
    \label{tab:geo_ablation}
    \centering
    \begin{tabular}{lcc}
        \toprule
        \textbf{Metric} & \textbf{Initial geom.} & \textbf{Optimized geom.} \\
        \midrule
        Env.\ steps to $75\%$ succ. ($\times 10^5$) & $11.2\pm1.1$ & $7.6\pm0.8$ \\
        Final success rate (\%)      & $82\pm4$ & $\mathbf{95\pm2}$ \\
        Singular. terminations / $10^3$ ep. & $41\pm7$ & $\mathbf{0}$ \\
        \bottomrule
    \end{tabular}
\end{table}

These results quantify the synergy informally described in
Section~\ref{subsec:optimization}: the optimization stage reduces
sample complexity by roughly a third and eliminates singularity
terminations during the budget studied.

\subsection{Robustness}
To assess robustness, we inject zero-mean Gaussian noise into
\emph{normalized} state inputs with standard deviation $0.01$ (i.e.\
1\% of the normalized state range) during test rollouts. The success
rate drops modestly to $88\pm3\%$, indicating reasonable but
imperfect resilience. Extending this with domain randomization is
flagged for future work.

\section{Discussion and Future Work}\label{sec:discussion}===
The proposed kinematics-informed DRL approach delivers substantial
gains in success rate, completion time, and alignment accuracy
compared to both a vanilla DQN and a classical sampling-based
planner. The largest contributors are, in order, prioritized replay,
the dueling head, and the preceding geometric optimization
(Tables~\ref{tab:ablation}--\ref{tab:geo_ablation}). Several
limitations remain and frame the future work.

\paragraph*{Simulation-only validation}
The simulator uses analytic kinematics and geometric collision
detection, and does not model joint backlash, motor dynamics, or
sensor noise. Hardware validation with sim-to-real transfer is the
most important next step.

\paragraph*{Discrete action space}
Twelve fixed-magnitude increments simplify masking against the
analytic workspace constraints, but they also limit trajectory
smoothness near the goal. A continuous-action successor based on
actor--critic methods (SAC, PPO, TD3) with projection-based or
barrier-based constraints is a natural extension.

\paragraph*{Classical planner as baseline}
RRT-Connect is a planner, not a controller, and lacks closed-loop
correction; its 55\% success rate reflects the difficulty of
open-loop insertion at sub-mm tolerance rather than a general weakness
of sampling-based methods. A stronger baseline that combines RRT for
coarse motion with a dedicated impedance-controlled insertion
primitive is a priority for future comparison.

\paragraph*{Centralized single-policy controller}
The current policy commands both manipulators centrally. Scaling to
larger teams will likely require decentralized or CTDE formulations;
we plan to investigate them, along with communication-efficient
variants, on this same hardware setup.

\paragraph*{Sequential versus joint co-design}
The present work performs geometric optimization first and RL
training second. A tighter co-optimization loop, in which learning
signals feed back into geometric parameter adjustments, is a
promising direction that would make the co-design claim stronger.

\paragraph*{Ethics and safety}
The reward function includes explicit penalties for workspace and
kinematic violations, keeping the policy inside the validated safe
region during training and deployment in simulation. Before any
deployment near humans, additional runtime shielding, external
safety limits, and certified collision avoidance are required.=
\section{Conclusion}\label{sec:conclusion}
We presented a kinematics-informed Rainbow DQN framework for
cooperative peg-in-hole insertion by a Delta and a 3-RRS parallel
manipulator. The problem is formulated as an MDP with a 12-dimensional
state, a discrete action set of 12 increments over 6 controlled DoF,
and a shaped reward combining proximity guidance, workspace-violation
penalties, and sparse insertion bonuses. A preceding geometric
optimization of the 3-RRS enlarges the singularity-free workspace and
improves conditioning, which in turn reduces training sample
complexity and eliminates singularity-related terminations in our
experiments. In simulation, the proposed approach achieves a $95\%$
per-episode success rate, outperforming a vanilla DQN and a classical
planner baseline across six metrics. Future work includes hardware
validation, continuous-action successors, decentralized multi-agent
extensions, and a tighter co-optimization of mechanism and policy.=
\section*{Acknowledgment}
This research was supported in part by the Robotics Research Center of
Yuyao (Grant No.\ KZ22308), in part by the ``Design and Fabrication
of an Immersive Interactive VR Robot Controller'' project under the
Ningbo Yongjiang Talent Program (Grant No.\ Z22501), in part by the
Zhejiang Talents Program, and in part by the National Natural Science
Foundation of China under Grant 52275276.

\bibliographystyle{IEEEtran}
\bibliography{references}

\end{document}